# Potato Leaf Disease Classification using Deep Learning: A Convolutional Neural Network Approach

Utkarsh Yashwant Tambe[1,*]
*Department of Data Science and Business Systems*
*SRM Institute of Science and Technology*
Kattankulathur (Chennai), India
utkarsh.tambe33@gmail.com

Dr. A. Shobanadevi[2]
*Department of Data Science and Business Systems*
*SRM Institute of Science and Technology*
Kattankulathur (Chennai), India
shobanaa3@srmist.edu.in

Dr. A. Shanthini[3]
*Department of Data Science and Business Systems*
*SRM Institute of Science and Technology*
Kattankulathur (Chennai), India
shanthia@srmist.edu.in

Hsiu-Chun Hsu[4]
*Department of Information Management*
*National Chung Cheng University*
Chiayi County, Taiwan R.O.C.
ellenj1022@gmail.com

*Abstract* - In this study, a Convolutional Neural Network (CNN) is used to classify potato leaf illnesses using Deep Learning. The suggested approach entails preprocessing the leaf image data, training a CNN model on that data, and assessing the model's success on a test set. The experimental findings show that the CNN model, with an overall accuracy of 99.1%, is highly accurate in identifying two kinds of potato leaf diseases, including Early Blight, Late Blight, and Healthy. The suggested method may offer a trustworthy and effective remedy for identifying potato diseases, which is essential for maintaining food security and minimizing financial losses in agriculture. The model can accurately recognize the various disease types even when there are severe infections present. This work highlights the potential of deep learning methods for categorizing potato diseases, which can help with effective and automated disease management in potato farming.

## I. INTRODUCTION

**Potato** is a vital food crop globally and a significant source of income for farmers. However, potato crops are susceptible to various diseases that can cause significant yield and quality losses. Leaf diseases are among the most common and destructive of these diseases [1]. Early detection and accurate diagnosis of these diseases are essential for effective disease management and control. Traditional methods of disease diagnosis rely on visual inspections, which are subjective and time-consuming. In recent years, there has been a growing interest in the use of deep learning techniques for automatic disease detection and classification.

The potato, **Solatium tuberosum**, as seen in fig. 1 is the fourth-ranked food crop used to feed an increasing global population because of its adaptability in cultivars and high complex carbohydrate content. For the dining and the processing market, potatoes can be kept for a long time, but storage disease issues are common [2]. Wherever potatoes are produced, diseases both in the field and during storage can be a limiting factor in the ability to produce them sustainably and profitably.

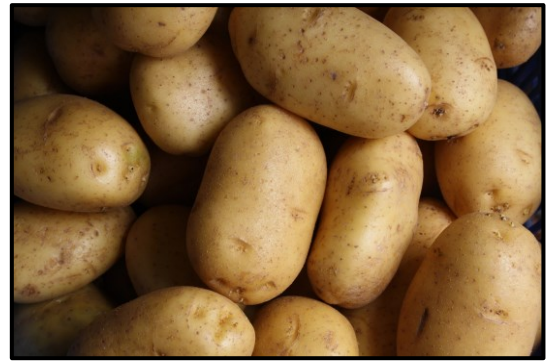

Figure 1: Potato (**Solatium Tuberosum**). Adapted from [3]

In this study, we suggest a deep learning method for categorizing potato leaf illnesses. In particular, we investigate the application of convolutional neural networks (CNNs) for the automated detection and classification of various potato leaf diseases. The primary goal is to develop a precise and useful method for the early detection and classification of potato leaf diseases so that farmers can act quickly and control the disease. The suggested method has the potential to greatly increase the effectiveness and accuracy of managing potato leaf disease and increasing crop yields while minimizing financial losses.

As, due to its capacity to extract intricate patterns and features from sizable databases, deep learning has become increasingly popular in recent years. Deep neural networks known as convolutional neural networks (CNNs) are frequently employed for picture classification and recognition tasks. They can learn hierarchical representations of images, which makes them especially suitable for this job.

It is built with the intention of detecting or identifying the various diseases that can affect potato leaves, because Convolutional Neural Network (CNN) can identify them with ease while the human eye cannot. Unbelievably, some pre-trained Neural Network Architectures have error rates of about **3%**, which is even lower than the best **5%** error rate of human vision. **5.1%** was determined to be the human top-5 error on large-scale images, which is higher than pre-trained networks [4].

## II. LITERATURE SURVEY

Singh et al. (2015) [5] - performed a review of the literature on various image processing techniques for detecting leaf disease. The writers wanted to speed up identification and detection of plant diseases while lowering the subjectivity that comes with naked-eye observation. They presented an algorithm that uses picture segmentation to automatically identify and categorise plant leaf diseases. The impact of HSI, CIELAB, and YCbCr colour spaces on disease spot detection was examined by the authors. To identify the disease spot, the Otsu technique was applied to the colour component after an image was smoothed using a median filter. The suggested method was not put to the test on any datasets.

He et al. (2015) [6] - deep residual networks were discussed in relation to image recognition challenges. As shallow representations for image retrieval and classification, VLAD and Fisher Vector are cited in the paper's literature review as related concepts. The writers also covered the advantages of encoding residual vectors over original vectors for vector quantization. The Multigrid method was also mentioned as a method for solving partial differential equations (PDEs) by reformulating systems as subproblems at multiple scales, where each subproblem is accountable for the residual solution between a coarser and a finer scale. This method is used in low-level vision and computer graphics.

Islam et al. (2017) [7] - suggested a method to identify diseases from images of potato plant leaves by combining image processing and machine learning. The authors classified diseases using the 'Plant Village' database of openly accessible plant images. Using the suggested method, the research classified 300 images of diseases with a 95% accuracy. The paper also highlights how crucial contemporary phenotyping and plant disease detection are to assuring food security and sustainable agriculture.

Velmurugan and Renukadevi (2017) [8] - describe research on the creation of software for the automatic detection and classification of plant leaf diseases. Only a few methods were suggested for particular databases, such as satellite images, leaf sets, maps, faces, fingers, etc., the authors discovered after conducting a literature review. The research suggests a novel algorithm that, with 94% accuracy, can identify and categorise the diseases under study. The processing process consists of four major steps: colour transformation, masking of green pixels, segmentation, and computation of texture statistics from the SGDM matrices. The suggested method was tried on a database of about 500 plant leaves, and the outcomes support the robustness of the suggested algorithm.

Dasgupta et al. (2019) [9] - address the issue of food scarcity in developing countries, particularly in India, due to potato production being affected by diseases such as Early Blight and Late Blight. To tackle this problem, the authors propose the use of deep learning and transfer learning techniques to develop a model capable of accurately detecting these diseases in potato plants. The literature review section of the paper delves into the fungus Alternaria Solani that is responsible for the early Blight disease. The disease has a polycyclic life cycle and requires free water to germinate. Early Blight primarily causes the premature defoliation of potato plants.

Cecilia et al. (2019) [10] - share a research on the use of image processing methods and machine learning algorithms for spotting pests and diseases in blueberry plants. Since there was no openly available database for this kind of fruit, the authors created their own database of images. The research used a variety of filters, including addWeighted and gaussianBlur to enhance the details in the images and medianBlur and gaussianBlur to remove noise. Utilizing both normalised and unnormalized versions of algorithms like HOG and LBP, traits were extracted. Using Deep Learning, the findings revealed an accuracy index of 84%. The report does not, however, contain a literature review.

Iqbal and Talukder (2020) [11] - investigated the use of image processing and machine learning methods in identifying and classifying diseases of potato leaf. In order to find diseased regions, the authors processed 450 images of healthy and diseased potato leaves from the freely accessible Plant Village database using image segmentation methods. To differentiate between the sick and healthy leaves, they used seven classifier algorithms, with the Random Forest classifier having the best accuracy (97%) rate. The study also included a review of prior studies that used machine learning and image processing to identify plant diseases.

## III. PROPOSED METHODOLOGY

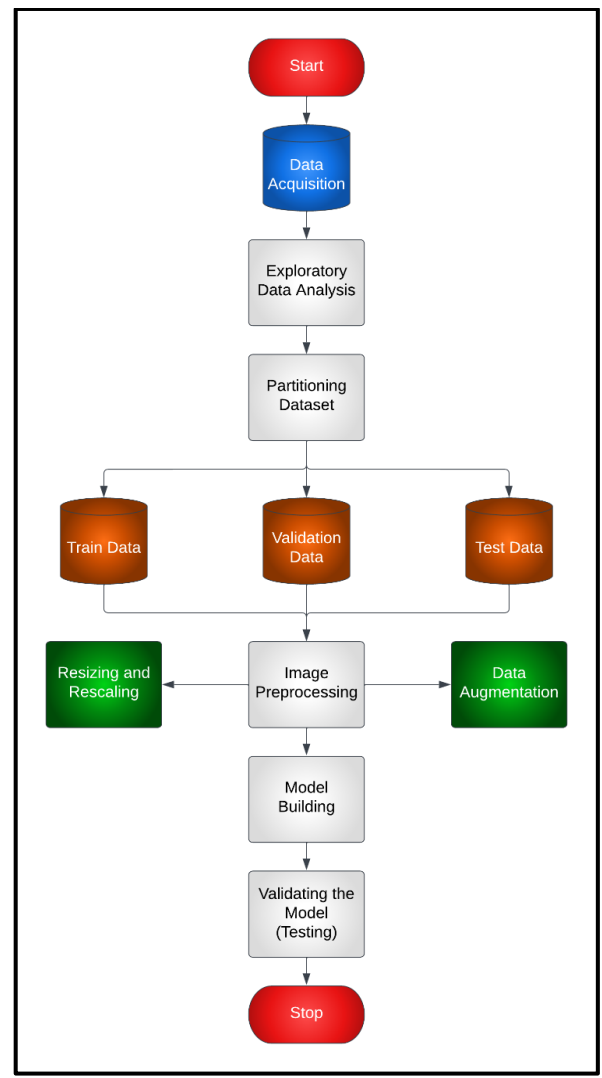

Figure 2: Methodology Flowchart

Data collection, pre-processing of the data, data augmentation, and disease classification make up the main four stages of the methodology described in this paper. Fig. 2 mentions the graphical representation of procedure for classifying different diseases of Potato leaf, step by step. So, the four mentioned steps are:

*A. Data Acquisition/Data Collection and Description*

Different image resolutions and sizes have been obtained from dataset uploaded on 'Kaggle', known as Plant Village by **Arjun Tejaswi** which contains plant diseases of around 15 types [12]. The dataset contains about 20,600 images of which 2152 images belong to plant 'Potato' and categorised as Healthy, Early-Blight, and Late-Blight as in fig. 3. The dataset is arranged in Joint Photographic Experts Group data format (.jpg file).

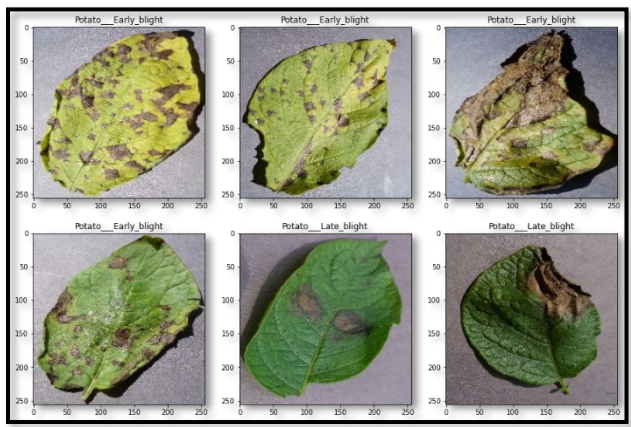

Figure 3: Images of Potato Leaves with different Diseases

*B. Data Preprocessing*

The potato-disease dataset is divided into three classes. These classes include Early-blight class (1000 images), Healthy class (152 images) and Late-blight class (1000 images). Hence, the dataset was divided into **46.4%**, **7.06%** and **46.4%** respectively.

The splitting of the dataset for training (70%), testing (15%) and validation (15%) is carried out. Before splitting the dataset, images of dataset were divided into batches for the ease of splitting the dataset. Also, specific functions like: 'prefetch' - loads next batch of images into CPU when GPU is busy training the images, 'cache' - saves time while reading the images; were applied for better use of resources of the system being used.

The CNN model is practiced on the train set. The validation set is used to fine-tune the hyperparameters (i.e., the design) of a classifier, while the test set provides an objective assessment of the final model fit on the training dataset [13]. Noise in the picture can first be reduced by removing any unnecessary parts. If there is too much disturbance in the image, it won't be used [14]. Pictures collected from various sources and of varying sizes must be resized to **256x256** pixels for the dataset's standardized input photos. The batch size is of **32**.

*C. Data Augmentation*

In contrast to the shallow networks used in machine learning, deep learning (Deep Network) needs a lot of data. Machine learning and deep learning frequently encounter issues with a lack of training data and an imbalance in the quantity of data for each class [15]. Data augmentation is the approach taken to solve this issue. Data augmentation is a technique for modifying data without changing its initial meaning. 5100 datasets are still insufficient for this research to achieve optimal performance, necessitating the use of data augmentation. The augmentation factors in this study are generated by the automatic application of simple geometric transformations like translations, rotations, scale changes, shearing, vertical and horizontal flips.

But for this dataset, only the rescaling and resizing of images (**1/255**), horizontal flip, vertical flip and random-rotation of **0.2** degrees is used as the leaves do not have high number of features to be considered.

*D. Disease Classification using CNN*

ML (Machine learning) in Artificial Intelligence (AI) includes DL (Deep Learning), also referred to as deep neural learning or deep neural network, have more levels than machine learning. Its state-of-the-art abilities in a variety of fields, including object detection, speech recognition, object categorization, and picture classification, have been enhanced by deep learning techniques [16]. Convolutional Neural Networks are one of the most popular deep learning choice. In some studies, CNN have been used to detect plant diseases based on the condition of the leaves [17]. CNNs, or Convolutional Neural Networks, usually have a few multi-layer convolutional layers that are arranged according to functions. One or more fully connected layers that are typical of a neural network commonly follow the subsampling layer. The input for the following feature layer is a feature collection that is contained on the preceding layer in a small area. CNN has a lot of expertise with computer vision issues. Compared to ANN (Artificial Neural Network), sharing parameters is one of its features, which lowers the number of parameters needed for the model. Additionally, CNN's features are of very high standard content. Softmax is the final output layer as we perform multiclass categorization. The sample model of CNN can be visualized in fig. 4 as given below.

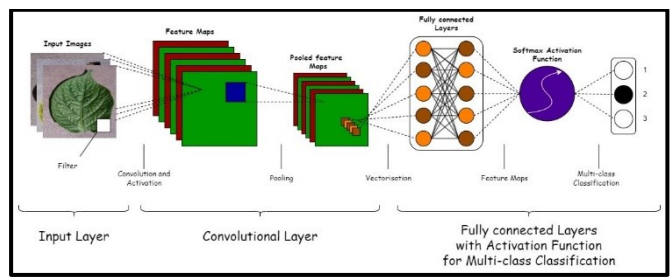

Figure 4: Convolution Neural Network (CNN)

In our research, we created a feature map by applying convolution to the input picture through a convolution filter. As a convolution filter, a 3x3 filter is employed. For this procedure, padding = "valid" is used. When padding is "valid," it must reflect zero padding. 5 convolution layers (C1, C2, C3, C4, C5) with a kernel/filter size of 3x3 and a subsampling layer (S1) with a max-pooling frame of 2x2 are included in the CNN model.

The input picture with varying sizes is convolved with 32 filters with no padding in the first convolution (C1), producing 32 features that are 254x254 in size. During the training process, the algorithms will learn these parameters. Rectified Linear Unit (reLU) are utilised as the activation function in each convolution layer. It makes the pictures' nonlinearity greater. These characteristics are produced by the C1 layer.

64 convolution filters are applied to the features map from the preceding layer in the second convolution layer (C2) to create 64 convolved features maps, each measuring 3x3. The second subsampling layer (S2), which employs max-pooling and a window size of 2x2, receives these features. As a result, 18496 shared feature maps, each measuring 125x125, are produced.

The third convolution layer (C3) uses 64 filters to create 64 feature maps, each measuring 3x3. The features are passed to S3, the third subsampling layer, which also employs max pooling with a window size of 2x2. The result is the creation of 36928 shared feature maps, each measuring 60x60.

There are 64 filters used in the fourth convolution layer (C4), producing 64 feature maps, each of which is 3x3. With a window size of 2x2, the fourth subsampling layer (S4), which employs maximum pooling, receives these generated features. Consequently, an image of 36928 pooled features with 14x14 pixels is produced.

64 filters are used to create 64 feature maps, each measuring 3x3 pixels, in the fifth and final convolution layer is (C5). These features are passed to the fifth subsampling layer (S5), which also employs maximum pooling and a window size of 2x2, using these features. The result is the creation of 36928 shared feature maps, each measuring 3x3. So, no dropout layer is employed. Dropout is a regularisation method that lowers overfitting. Only the model's training step uses dropout. Dropout disables some randomly selected neurons' capacity to activate downstream neurons on the forward pass and does not apply any weight updates to the neurons during the backward run. This ignores the neurons during the training phase in line with the dropout rate.

The model adds a flatten layer as it flattens the data into a 1-dimensional array for entry into the following layer [18]. Following that, a dense layer with 64 nodes receives the produced features. A dense layer with 64 nodes feeds these generated characteristics. There are 1,47,520 parameters total here. Finally, for the classification, we used a second dense layer with softmax activation and 3 nodes. Thus, there will be 195 factors in total.

The algorithm has a total of 277,891 trainable factors and the same are shown in table 1, below.

TABLE I
DEPICTION OF LAYERS AND ITS PARAMETERS INVOLVED DURING TRAINING THE MODEL

| Layers | Parameters |
| --- | --- |
| Conv2D (C1) | 896 |
| Conv2D (C2) | 18496 |
| Conv2D (C3) | 36928 |
| Conv2D (C4) | 36928 |
| Conv2D (C5) | 36928 |
| Dense (1) | 147520 |
| Dense (2) | 195 |
| **Total** | **277891** |

## IV. RESULTS AND DISCUSSIONS

### A. Training Process

The CNN model is finally created, and then the **Adam** optimizer is used to build the model. It is a highly effective optimisation method used for deep neural network training. It incorporates the benefits of two optimisation techniques, RMSProp (Root Mean Square Propagation) and AdaGrad (Adaptive Gradient Algorithm). The five-layer Convolutional Neural Network architecture model was used for the dataset learning exercise. The study's period specification called for 50 epochs, a 32-batch size, and a learning rate of 0.01 to enhance the model's performance. The suggested method's algorithm searches for values in the image dataset as part of the learning process in order to recognise new images. Each epoch phase's value must correspond to the value of the picture being trained. To calculate the loss and accuracy values, the outcomes of the epochs are recorded. The accuracy value indicates how accurately the system classifies objects, and the value of the loss gained must be close to or equal to zero. Loss is an indication that the model has a poor value.

In both plots shown in fig. 5, the findings of the Accuracy and Loss values are displayed. obtaining values that are near to one and other to zero. While training, the early and later epochs show more fluctuation values. After the 50th epoch, the training dataset's accuracy, which includes the original images, achieves **98.189%**. The CNN needed 7000 seconds, or approximately 1.9 hours, to complete 50 epochs at an average speed of 104 seconds per epoch.

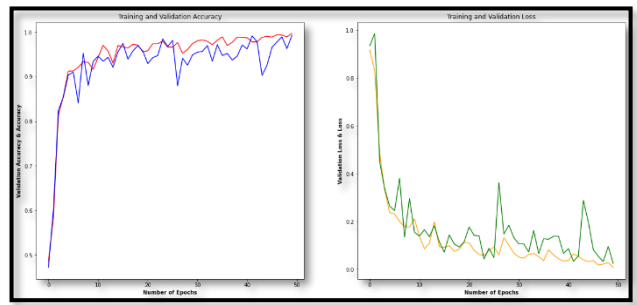

Figure 5: Accuracy and Loss recorded while Training the Dataset

### B. Testing Process

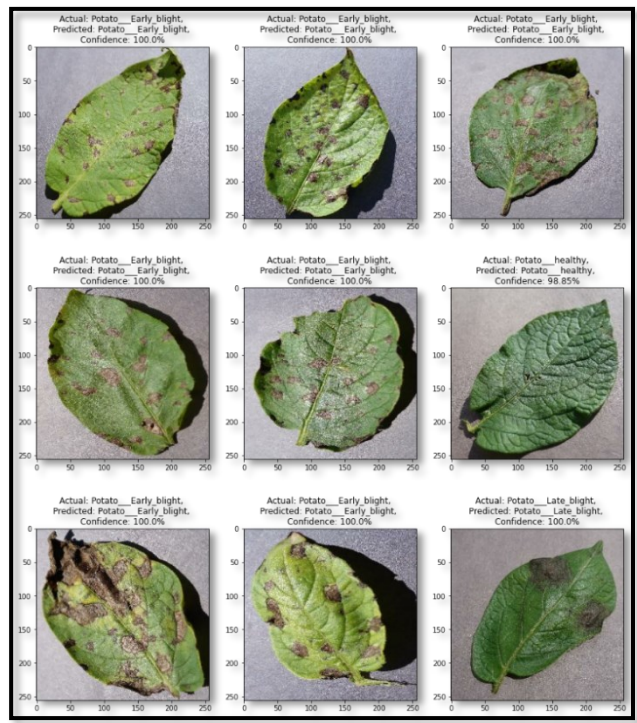

Figure 6: Testing the Trained Model on Test Dataset for Accurate Classification of Disease

Following the training procedure using the generated dataset, we retrieve the remaining part of the dataset i.e., testing dataset to be tested with the model. The testing accuracy is recorded as **99.18%** and testing loss as **2.48%**. Refer, to the fig. 6 for results.

Also, the model trained and tested is **capable of signifying** the **confidence** up to what extent it is able to classify or predict the disease.

When assessing the model's accuracy with a leaf image object, it is crucial to keep in mind that image noise will result in subpar classification results. By cutting out everything, but the desired leaf object and ensuring that there is only one leaf in each frame, noise can be removed from a picture.

## V. Conclusion and Future Work

In this research, we suggested a deep learning strategy for categorising potato leaf diseases using a CNN (Convolutional Neural Network). Pre-processing the leaf images, training a CNN model on the pre-processed data, and assessing the model's performance on a test set are all steps in the suggested approach. With an overall accuracy of **99.18%**, the experimental findings showed that the CNN model was highly accurate in classifying two different potato leaf diseases, including Early Blight and Late Blight.

The suggested method may offer a trustworthy and effective means of diagnosing potato disease, which is essential for ensuring food security and minimising financial losses in agriculture. This approach can be further extended to other plant species, enabling early detection of diseases, and facilitating prompt interventions to prevent further spread.

Overall, the findings of this research show the efficacy of deep learning-based methods for identifying plant diseases and their potential to revolutionise the agricultural technology industry [19]. The proposed approach can assist farmers and experts in the early detection of diseases and timely implementation of control measures, thereby improving crop yields, reducing losses, and contributing to global food security.

Furthermore, the proposed deep learning-based approach offers several advantages over traditional methods of plant disease diagnosis. Unlike manual inspection, which is time-consuming and prone to human error, deep learning-based approaches can provide accurate and consistent results with minimal human intervention. Additionally, this approach can process large amounts of data in a short amount of time, making it ideal for real-time monitoring and detection of plant diseases.

While the proposed approach achieved high accuracy in identifying potato leaf diseases, further research is necessary to investigate its performance on other plant species and under different environmental conditions. Additionally, the proposed method can be further improved by incorporating additional data sources and developing more efficient models on these datasets.

In conclusion, this study demonstrates the potential of deep learning-based approaches for plant disease diagnosis and management. Early detection of potato leaf diseases is made possible by the suggested technique, which can have a significant impact on crop protection and global food security. With further research and development, this approach can be extended to other crops and environmental conditions, enabling the development of more effective and sustainable agricultural practices.